\documentclass{article}
\usepackage[ruled,linesnumbered,noresetcount,vlined]{algorithm2e}
\makeatletter

\newcommand{\nosemic}{\renewcommand{\@endalgocfline}{\relax}}
\newcommand{\dosemic}{\renewcommand{\@endalgocfline}{\algocf@endline}}
\let\oldnl\nl
\newcommand{\nonl}{\renewcommand{\nl}{\let\nl\oldnl}}

\SetKwFor{Repeat}{repeat}{:}{endw}
\makeatother
\usepackage[nonatbib,final]{neurips_2021}
\usepackage[numbers]{natbib}
\usepackage[utf8]{inputenc} 
\usepackage[T1]{fontenc}    
\usepackage{hyperref}       
\usepackage{url}            
\usepackage{booktabs}       
\usepackage{amsfonts}       
\usepackage{nicefrac}       
\usepackage{microtype}      

\usepackage{amsmath,mathrsfs,amssymb,mathtools,bm}
\usepackage{scalerel}
\usepackage{amsthm}
\usepackage{txfonts}
\usepackage{fontawesome}
\usepackage{wrapfig}
\usepackage{todonotes}
\usepackage{xcolor}
\usepackage{setspace}
\usepackage{bm,bbm}
\usepackage{paralist}
\usepackage{enumitem}
\usepackage{subcaption}

\makeatletter
\newcommand{\oset}[3][0ex]{%
  \mathrel{\mathop{#3}\limits^{
    \vbox to#1{\kern-2\ex@
    \hbox{$\scriptstyle#2$}\vss}}}}
\makeatother

\usepackage{letltxmacro}
\LetLtxMacro\orgvdots\vdots
\LetLtxMacro\orgddots\ddots

\newtheorem{problem*}{Problem}

\newtheorem{theorem}{Theorem}

\newtheorem{definition}{Definition}

\newtheorem*{example*}{Example}

\usepackage{multirow,mdframed}
\mdfsetup{skipabove=0pt,skipbelow=0pt}
\mdfdefinestyle{model}{%
  innertopmargin=0pt,
  innerbottommargin=0pt,
  innerleftmargin=0pt,
  innerrightmargin=0pt,
  skipbelow=0pt,%
  skipabove=0pt,
  splitbottomskip=0pt,
  splittopskip=0pt,
  leftmargin =0pt,%
    rightmargin=0pt,
  splittopskip=0pt,
    usetwoside=false,
}
\usepackage{float}
\floatstyle{ruled}
\newfloat{model}{thp}{lop}
\floatname{model}{Model}

 \newcommand{\cL}{\mathcal{L}}
\newcommand{\cM}{\mathcal{M}}

\usepackage{esvect}

\title{Context-Aware Differential Privacy for Language Modeling}

\author{
  My H.~Dinh\\
  Syracuse University\\
  \texttt{mydinh@syr.edu}\\
  \And
  Ferdinando Fioretto\\
  Syracuse University\\
  \texttt{ffiorett@syr.edu}\\
}

\begin{document}

\maketitle\sloppy\allowdisplaybreaks

\begin{abstract}

The remarkable ability of language models (LMs) has also brought challenges at the interface of AI and security. A critical challenge pertains to how much information these models retain and leak about the training data. This is particularly urgent as the typical development of LMs relies on huge, often highly sensitive data, such as emails and chat logs.
To contrast this shortcoming, this paper introduces Context-Aware Differentially Private Language Model (CADP-LM) , a privacy-preserving LM framework that relies on two key insights: First, it utilizes the notion of \emph{context} to define and audit the potentially sensitive information. Second, it adopts the notion of Differential Privacy to protect sensitive information and characterize the privacy leakage. A unique characteristic of CADP-LM is its ability to target the protection of sensitive sentences and contexts only, providing a highly accurate private model.  Experiments on a variety of datasets and settings demonstrate these strengths of CADP-LM.

\end{abstract}

\section{Introduction}

Language models (LMs) are essential components of state-of-the-art natural language processing. 
Their recent development has focused on training increasingly large models, containing hundreds of millions parameters, giving rise to a new generation of tools with remarkable abilities in sentence completion, code generation, text-to-image translation, and reasoning, to mention only a few examples \cite{gpt3, radford2019language}. 
To obtain these remarkable performances, LMs are routinely trained on huge and often highly sensitive datasets, such as emails, chat logs, and personal text. However, the size of these models combined with the sensitive nature of the training data creates a dangerous mix: It is now well documented that LMs memorize and regurgitate large potions of their training data \cite{Carlini2019, carlini2020, inan, vitaly}. 
The tendency to memorize data can lead to a leakage of sensitive data from a model’s training set, where the behavior of the model on samples that were present in the training set becomes distinguishable from samples that were not. These privacy concerns are critical and can cause profound damage to both data users and data curators. For example, by querying a model trained on patient record data, an adversary could guess with high confidence if an individual contributed to the training set, or could recover sensitive information of the individuals having some medical condition.

To address these concerns, recent work has focused on developing privacy-preserving language models \cite{kerrigan-etal-2020-differentially, Ramaswamy}. In particular, Differential Privacy (DP) \cite{dwork:14} has become the paradigm of choice for protecting data privacy. In the context of machine learning, DP ensures that algorithms can learn the relations between data and predictions while preventing them from memorizing sensitive information about any specific individual in the training data.
While this property is appealing, the application of DP to large LMs is challenged by the resulting poor model utility or even non convergence issues \cite{kerrigan-etal-2020-differentially}. 
DP uses carefully calibrated noise to render the models' outputs insensitive from the contributions of each individual sample. The application of traditional DP to a training process, however, considers the protection of samples as a whole. Thus, it induces protection for those record's attributes which are not sensitive, resulting in overly pessimistic protection algorithms. Moreover, current implementations of DP mechanisms for LMs can only provide protection guarantees for individuals' records when they have a clearly defined format or structure. They ignore the fact that the same piece of private information can be represented in different ways. For example, one's social security number can be expressed as a mix of words and numbers.

On the other hand, privacy preserving mechanisms for language models can be more effective when they are aware of contextual information which can reveal sensitive data \cite{brown}. This work mitigates these issues by introducing Context Aware Differentially Private Language Model (CADP-LM), a privacy-preserving LM mechanism which relies on a modified notion of Differential Privacy that focuses on the protection of sensitive attributes only. CADP-LM is motivated by the intuition that direct protection of sensitive tokens in language data may be insufficient. Since contextual information preceding the token may be used to recover the token itself, an effective privacy preservation strategy must also consider the identification and protection secret-revealing context.

In summary this work {\bf (1)} Introduces the notion of context for language model and connect it with the development of privacy preserving frameworks, 
{\bf (2)} it provides a theoretical privacy notion for DP language model with context, and 
{\bf (3)} it experimentally demonstrates the ability of CADP-LM to contrast privacy attacks and retain high accuracy.

\section{Language Models and Privacy Risks}

\paragraph{Language model.}
A language model places a probability distribution $p(\bm{x})$ over discrete token sequences $\bm{x} = (x_1, x_2, \ldots)$.
Learning such distribution is achieved by a chain rule factorization and modeling the conditional distribution over a single \emph{target token} given a \emph{context} of previous tokens:
\[
    p(\bm{x}) = \prod_{i=1^n} p(x_i | x_1, \ldots, x_{i-1}).
\]
Given a corpus $D =\{\bm{x}^1,\ldots \bm{x}^N\}$, with $N = |D|$, the learning task trains a neural network parametrized by  $\theta$ to learn $p(\bm{x})$ by minimizing the negative log-likelihood over $D$:
\[ 
 \cL(D) = -\sum_{t=1}^{N} \sum_{i=1}^{n_t} \log p_\theta(x_i^t| \bm{x}_{1}^t, \ldots \bm{x}_{i-1}^t)
\]


 
The negative log-likelihood of a sequence $\bm{x}$ is also referred to as \emph{perplexity} to describe how "surprised" is the model to see a given token. Low perplexity scores are associated with confident model predictions. This is also often used as a proxy to quantify how likely has the model been to see a specific token during training. 

\paragraph{Threat model and attacks to LMs.}
This paper considers a blackbox access to the model in which the attacker observes the model's output probabilities (or logits)  in response to a given query. The attacker queries the model via seeding prompts, and their querying ability is considered \emph{unrestricted}. 

We now review two popular privacy attacks against LMs. These attacks will later be used to evaluate the effectiveness of privacy preserving mechanism developed in this paper. 

\smallskip\noindent $\bullet$ \textbf{Canary insertion.} 
Proposed by Carlini et al.~\cite{Carlini2019} this attack recovers the original training data point using only query prompts. The assessment of this attack works by inserting \emph{canaries} (e.g., random sequences) into the training dataset, and then measuring if the model has unintentionally memorized such canaries using 
an exposure metric function: 
\[
    \mathrm{exposure}_{\theta}(s[r])   = \log_2 |\mathcal{R}| - \log_2 \mathrm{rank}_{\theta}(s[r])  
\]
 where $s[r]$ is a canary, $\theta$ is a vector of parameters for the LM model, $\mathcal{R}$ is an event set, and  
 \(
 \mathrm{rank}_{\theta}(s[r]) = | r' \in \mathcal{R} : p_\theta(s[r']) 
    \leq p_\theta(s[r]) |
 \) 
 is the index of $s[r]$ in the list of all possibly-instantiated canaries, ordered by the empirical model perplexity of all those sequences.
 
For example, a canary takes the format $s =  \textsl{"My SSN is  XXX"}$, where $\textsl{XXX}$ is filled with random values $r$ from event $\mathcal{R} = [9]^3$. 
A low exposure value for that canary indicates a low risk of its leakage from the model. 

\smallskip\noindent $\bullet$ \textbf{Membership inference.} 
An attacker performs a membership inference when they try to ascertain the membership of a given sequence to the training dataset \cite{Carlini2019}. This is achieved by simply querying the model's perplexity scores for some prompts, ranking them, and choosing the ones with the lowest perplexity, i.e., highest likelihood they appear in the training set. 


\section{Differentially Private Machine Learning}

Differential privacy (DP) \cite{dwork:14} is a strong privacy notion used to quantify and bound the privacy loss of an individual's participation in a computation. A differentially private machine learning model bounds the amount of knowledge an attacker may collect (by observing the model's outputs) about membership of an individual's data into the training set.
The action of adding or removing a record from a dataset $D$, resulting in a new dataset $D'$, defines the notion of \emph{adjacency}, denoted $D \sim D'$.
\begin{definition}
  \label{dp-def}
  A mechanism $\cM \!:\! \mathcal{D} \!\to\! \mathcal{R}$ with domain $\mathcal{D}$ and range $\mathcal{R}$ is $(\epsilon, \delta)$-differentially private, if, for any two adjacent datasets $D \sim D' \!\in\! \mathcal{D}$, and any subset of output responses $R \subseteq \mathcal{R}$:
  \[
      \Pr[\cM(D) \in R ] \leq  e^{\epsilon} 
      \Pr[\cM(D') \in R ] + \delta.
  \]
\end{definition}
\noindent 
Parameter $\epsilon > 0$ describes the \emph{privacy loss} of the algorithm, with values close to $0$ denoting strong privacy, while parameter 
$\delta \in [0,1)$ captures the probability of failure of the algorithm to satisfy $\epsilon$-DP.

\smallskip\noindent{\textbf{DP-SGD}.} 
DP-Stochastic Gradient Descent \cite{abadi:16} (DP-SGD) is arguably the most commonly adopted DP ML algorithm. In a nutshell, DP-SDG computes the gradients for each data sample in a random mini-batch, clips their $L_2$-norm, adds noise to ensure privacy, and computes their average. The privacy loss is tracked at each training iteration using the moment accountant \cite{abadi:16}. 


\section{Challenges of Privacy-Preserving DP}

Two well documented issues of DP-SGD for language modeling tasks are the deterioration in performances and poor convergence \cite{kerrigan-etal-2020-differentially, bagdasaryan}. The issues are associated with the noise injection step that scales with the number of parameters, resulting in poor gradients updates on large LMs. 

These challenges can be partially overcome if one could identify the sensitive information in the training set and adopt a DP strategy to protect exclusively such information. This identification module can be derived from pre-existing privacy policies, or off-the-shelf Name Entity Recognition (NER) models, as explored in \cite{shi-etal-2022-selective}. While potentially effective, however, these methods make assumptions about the structure of data to be protected. For example, they may exploit the fact that credit card numbers are usually described as strings of 12 digits, emails have an "@domain." string, etc. However, structural assumptions fail when attempting to protect sensitive information more generally \cite{brown}. 
More importantly, a hand-crafted method to recognize sensitive tokens ignores the \emph{context} in which the private information is shared. This concept is important because sensitive information may be described in multiple, often ambiguous, ways and context is often useful to infer whether an information may be sensitive. For example, consider a partial sequence ``my number is''; While the next token may have a structure conformant with a pre-existing secret policy, this information may or may not be sensitive, based on the context. If the extended sentence was ``I received my social security card; My number is'', then the next token is likely to be sensitive, while if the sentence was ``I  am waiting at the motor vehicle department; My number is``, then this information may not be sensitive.
\emph{This paper argues that the context in which the sensitive information is shared is as important as the sensitive information itself.} 

\section{Context-Aware Differentially Private Language Models}
\label{sec:alg}

To address these limitation, we introduce context-aware DP language model (CADP-LM), which actively detect the context in which sensitive information may be revealed.

We start by introducing the notion of \emph{context} more formally. It considers the presence of a semantic invariant mapping $\phi$ which transforms sequences into other sequences with similar semantics. The function $\phi$ can be a mapping from a sentence to another which uses synonymous words, or a text summarization procedure, or a back-translation language module.

\begin{definition}
\label{context}
Consider a text sequence $\bm{x} = (x_1, \ldots, x_{i-1})$ and let $x_i$ being the (possibly sensitive) token to be predicted. The $\alpha$-context of $x_{i}$ is the smallest subsequence $\tilde{\bm{x}}$ of $\bm{x}$ such that: 
\[
    \left| p \left(x_i \;\vert\; \bm{x} \right) 
         - p \left(x_i \;\vert\; \phi(\tilde{\bm{x}}) \right) \right| \leq \alpha.
\]
\end{definition}
This notion of (approximate) context will be useful to quantify the properties of a secret-triggering sequence, so that a model can be constructed to detect such sequences. In the above definition, $x_i$ plays the role of a secret token and the context is the portion of its preceding sequence that has similar ability to reveal the token itself.

At a high level CADP-LM is composed of two steps: 
\begin{enumerate}
  \item We first train a \emph{context-aware sensitive detection} 
  $\cM_\phi$ module whose goal is to recognize whether a sequence may contain a sensitive token. 

  \item Next, we apply such detection module to a training corpus, and apply DP-SGD only to the sequences detected as secret-triggering.
\end{enumerate}

Details of the algorithm and formal guarantees are discussed next.

\paragraph{Context-aware sensitive detection}
Given a training corpus, we train a context-aware sensitive detection model $\cM_\phi$ to distinguish sequences which are sensitive triggers from those which are non-sensitive.
As briefly hinted above, there is not a unique way to generate a training dataset for a context-aware sensitive detection model $\cM_\phi$ since there is not a unique way to define a semantic invariant function $\phi$. In this paper, we use a simple yet effective invariant
function which transforms a sequence $\bm{x} = (x_1, \ldots, x_{i-1})$ into $\phi(\bm{x})$ that satisfies the $\alpha$-context notion introduced above, using a round trip translation strategy \cite{jiang-etal-2021-know}. This translates the original context seeding prompt $\bm{x}$ into another language and back to attain the semantic invariant transformation. We use $\phi(x)$ to augment the original training set and train model $\cM_{\phi}$ recognize dangerous sequences generated from different prompt mining strategies.

The next module assumes the presence of such a detection model with true-positive rate $\gamma$. In other words, $\cM_\phi$ fails to predict whether sequence $\bm{x}$ is secrete triggering with probability $(1-\gamma)$.

\paragraph{Context-aware Differential Privacy}
The next step uses the context-aware sensitive detection module $\cM_\phi$ to partition the training corpus $D$ into a sensitive $D_S$ and non-sensitive $D_{NS}$ subsets. We also write $N_S = |D_S|$ and $N_{NS} = |D_{NS}|$ to denote the size of the sensitive and non-sensitive datasets. 

Consider a training corpus $D = (D_S, D_{NS})$ where $D_S$ is the subset of $D$ containing sensitive sequences, while $D_{NS}$ contains non-sensitive sequences. We are interested in protecting the sensitive information in $D_S$, therefore the notion of dataset adjacency of differential privacy defines the change of a single sequence in $D_S$ only. Thus, two datasets $D = (D_S, D_{NS})$ and $D' = (D_S', D_{NS})$ are adjacent if $D_S$ and $D_S'$ differ in at most a single entry.

CADP-LM is described in Algorithm \ref{alg:alg1}. For each sample of the training corpus $D$, the training procedure uses mechanism $\cM_\phi(\bm{x})$ to classify $\bm{x}$ as a sensitive or non-sensitive element. 
The algorithm uses (noisy) gradient updates on sensitive sentences (i.e., those predicted to be in $D_{S}$) using a DP-SGD step, and uses exact gradients for those samples predicted to be in $D_{NS}$. 



 
 \begin{algorithm}[H]
  \caption{CADP-LM\!\!\!\!\!\!\!\!\!\!\!\!\!\!\!\!}
  \label{alg:alg1}
  \setcounter{AlgoLine}{0}
  \SetKwInOut{Input}{input}

  \Input{Training corpus $D$ ; Context detection function $\psi$; 
    Iterations $T$;~Noise\!\!\!\!\!\!\!\!\!\!\!\!\!\!\!\!\\variance $\sigma^2$; Clipping bound $C$; Learning rate $\eta$} 
  \label{line:1a}
  \For{iteration $t =  1,2, \ldots T$} {    
  \label{line:2a}
    \ForEach{minibatch $B$ of $D$} {
    Construct $(B_S, B_{NS})$ by applying $\cM_\phi(\bm{x})$ on samples $\bm{x} \in B$
    
    \label{line:3a}
    Perform private updates with DP-SGD($\sigma^2, C, \eta)$ on $B_{S}$
    
    \label{line:4a}
    Perform regular updates with SGD($\eta$) on $B_{NS}$
    }
  }
 
\end{algorithm}
 
The following results consider a CADP-LM algorithm that trains a model over T epochs using a sensitive dataset $D_S$ detected by $\cM_\phi$ containing $N_s$ training samples, uses mini-batches $B$ at each iteration, and standard deviation parameters $\sigma$. 

\begin{theorem}
\label{main}
CADP-LM satisfies 
satisfies $\left(\frac{T\,N_S \varepsilon}{|B|} + \frac{\log(\nicefrac{1}{\delta})}{\alpha-1}, \delta\right)$-DP, for any $(1-\gamma) < \delta < 1$. 
\end{theorem}
The result is obtained by noting that each DP-SGD step satisfies $(\alpha, \varepsilon)$-Renyi differential privacy which also satisfies 
$(\varepsilon + \frac{\log(\nicefrac{1}{\delta})}{\alpha-1}, \delta)$-DP \cite{abadi:16}, for $0 < \delta < 1$. 
The DP parameter $\delta$ is lower bounded by $(1-\gamma)$, the failure probability to detected a sensitive sample in minmibatch $B$.

 
\section{Experimental Analysis}
This section first describes the experimental settings. It then illustrates the performance of the proposed CADP-LM over several metrics and compares it against other DP approaches to train LM. 

\subsection{Training Details}

\smallskip\noindent{\textbf{LM details.}} The LM architecture uses a single LSTM layer with an embedding size of 200 and a hidden layer of size of 200. 

\smallskip\noindent{\textbf{Context detection model $\cM_\phi$ details}.} 
The context detection model was constructed by fine-tuning \textsl{DistilBERT} \cite{https://doi.org/10.48550/arxiv.1910.01108}, as provided by Hugging Face. The task is a binary classification task over the dataset \textsl{WikiText-2}.
Standard canaries of format "My bank security code is" where inserted in the dataset and labeled as sensitive. All other sequences were labeled as non-sensitive. 
A T5 model \cite{https://doi.org/10.48550/arxiv.1910.10683} (from Hugging Face )  was selected as a semantic invariant mapping $\phi$ 
to perform a round-trip translation to 10 different languages.
The trained detection model was able to detect (with 100\% accuracy) the sensitive sentences containing sequence "My bank security code is" and very high accuracy (i.e., $>98\%$) for other variants of this sensitive canary, such as "My new bank security code is", "I went to the bank to get my new card whose code is", etc. Note that, when querying a non private LSTM model with such sequences we were able to retrieve the secret tokens.

\smallskip\noindent{\textbf{Baselines}.} The experiments adopt three baselines: \textsl{noDP}, a non-private language model, \textsl{DP-SGD} a language model trained using DP-SGD, and  \textsl{S-DPSGD} \cite{shi-etal-2022-selective}. The latter is a very recent proposal which also uses DP-SGD on subsets of a corpus. Differently from CADP-LM, however, S-DPSGD focuses on protecting secret tokens only using predefined format detection mechanisms, and thus ignoring the concept of context. 

\smallskip\noindent{\textbf{Dataset}.} The experiments use two datasets: \textsl{WikiText-2} \cite{merity2016pointer} and \textsl{Reddit clean jokes} \cite{Redditdataset}. An 80-20 split was used in both datasets.

\subsection{Attack Details}

\smallskip\noindent{\textbf{Canary insertion}.}
The canary insertion attacks were configured as follows. The canary "My bank security code is 450." was inserted in the training data 450 times in Wikitext-2 and 5 times in Reddit. The LM models were then trained with canary of format $s=$"My bank security code is XXX", where XXX is a random string from $\mathcal{R} = [9]^3$. The exposure scores are obtained from the predicted ranks of the inserted canaries. 

\smallskip\noindent{\textbf{Membership inference}.} Membership inference attacks where configured as follows. The membership inference attack dataset was created by randomly selecting 50 protected secrets from the training set and 50 samples from the test set. A random guess would give an accuracy of 50\%. Then, the top 50 lowest perplexity sequences are selected as training samples and the rest as testing samples.

\subsection{Results Discussion}  

\begin{figure}
     \centering
     \begin{subfigure}[b]{0.3\textwidth}
         \centering
         \includegraphics[width=\textwidth]{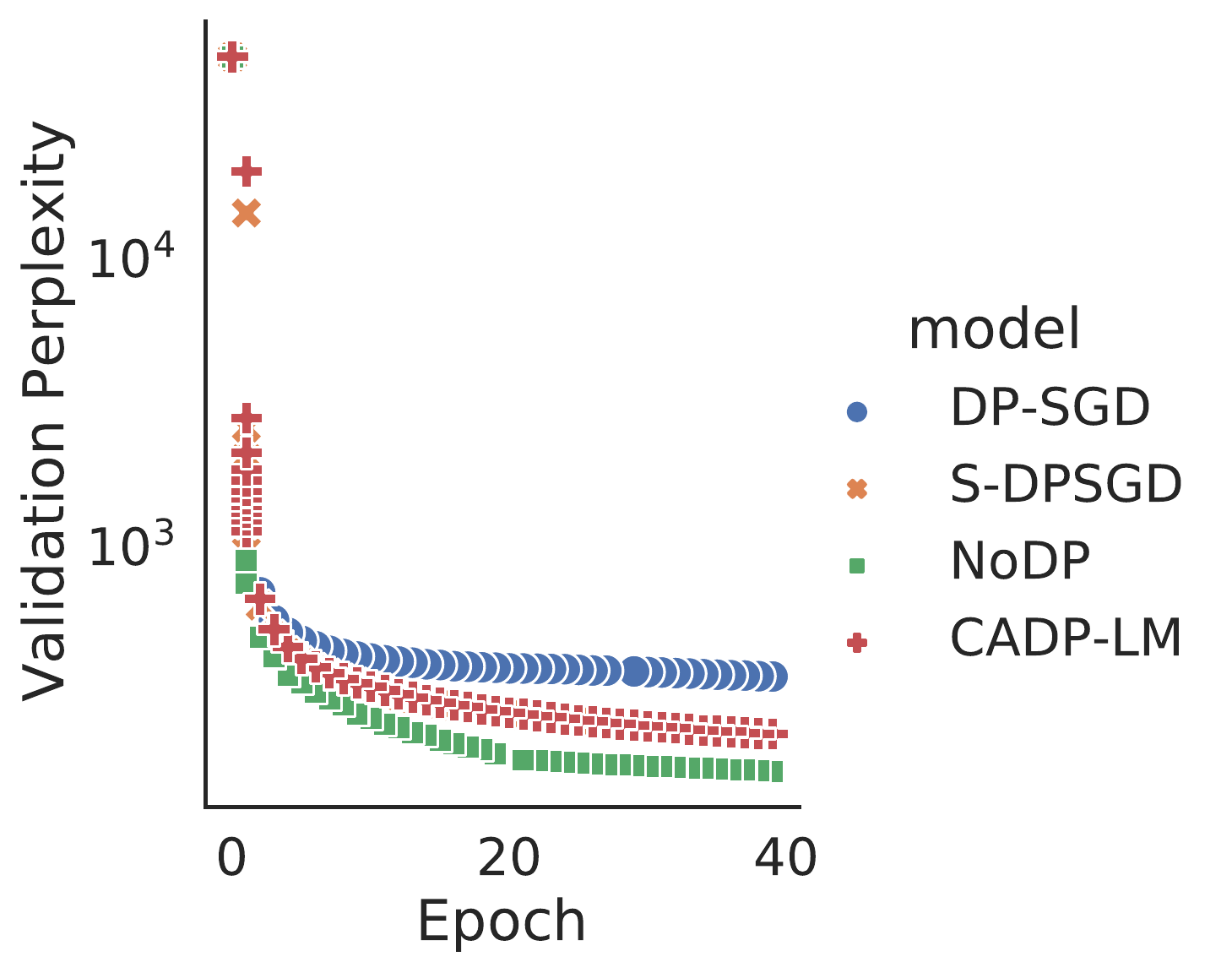}
         \caption{Learning curve}
        \label{fig:wikiacc}
     \end{subfigure}
     \hfill
     \begin{subfigure}[b]{0.3\textwidth}
         \centering
         \includegraphics[width=\textwidth]{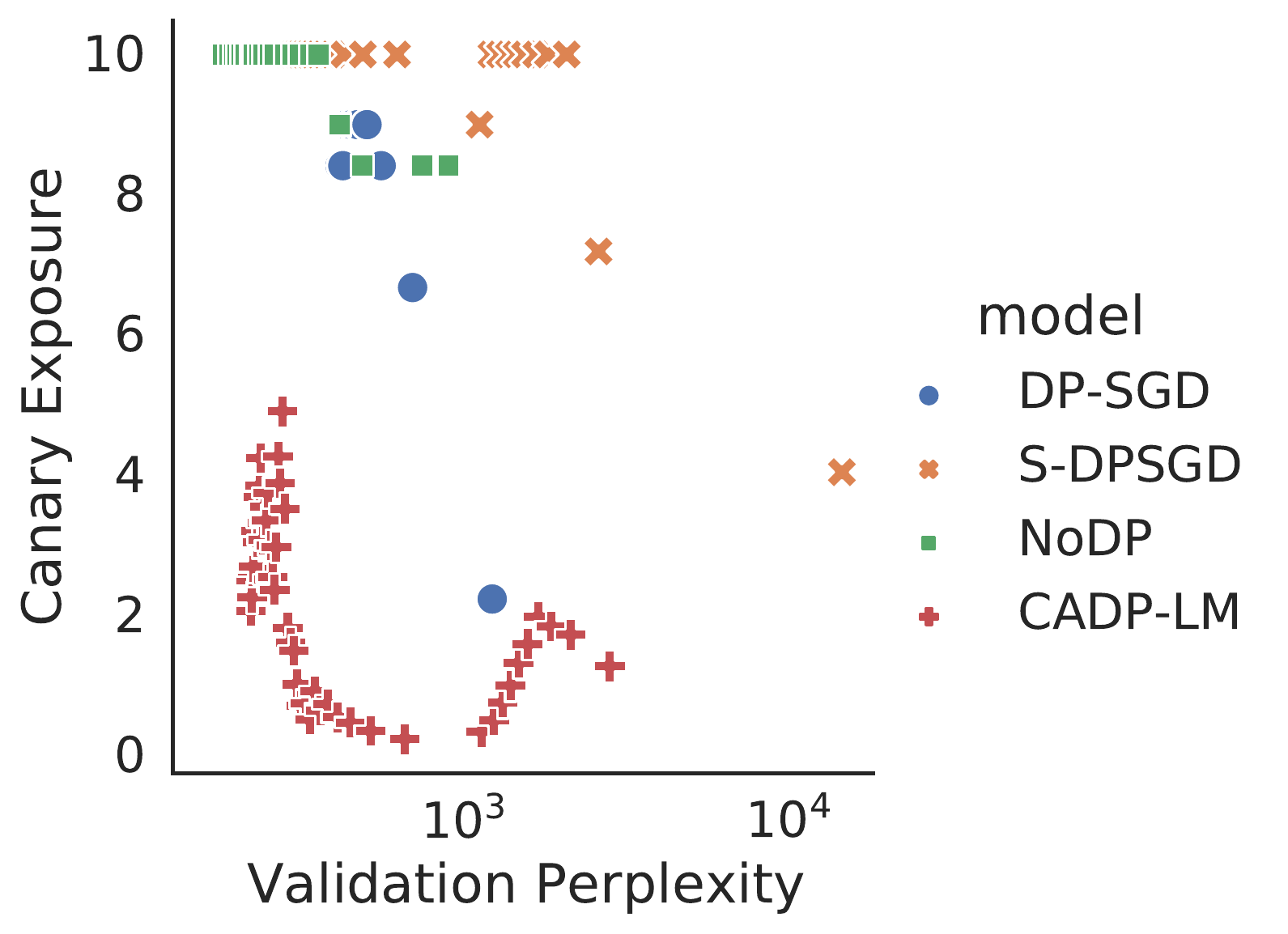}
         \caption{Canary Insertion}
         \label{fig:wiki_insert}
     \end{subfigure}
     \hfill
     \begin{subfigure}[b]{0.3\textwidth}
         \centering
         \includegraphics[width=\textwidth]{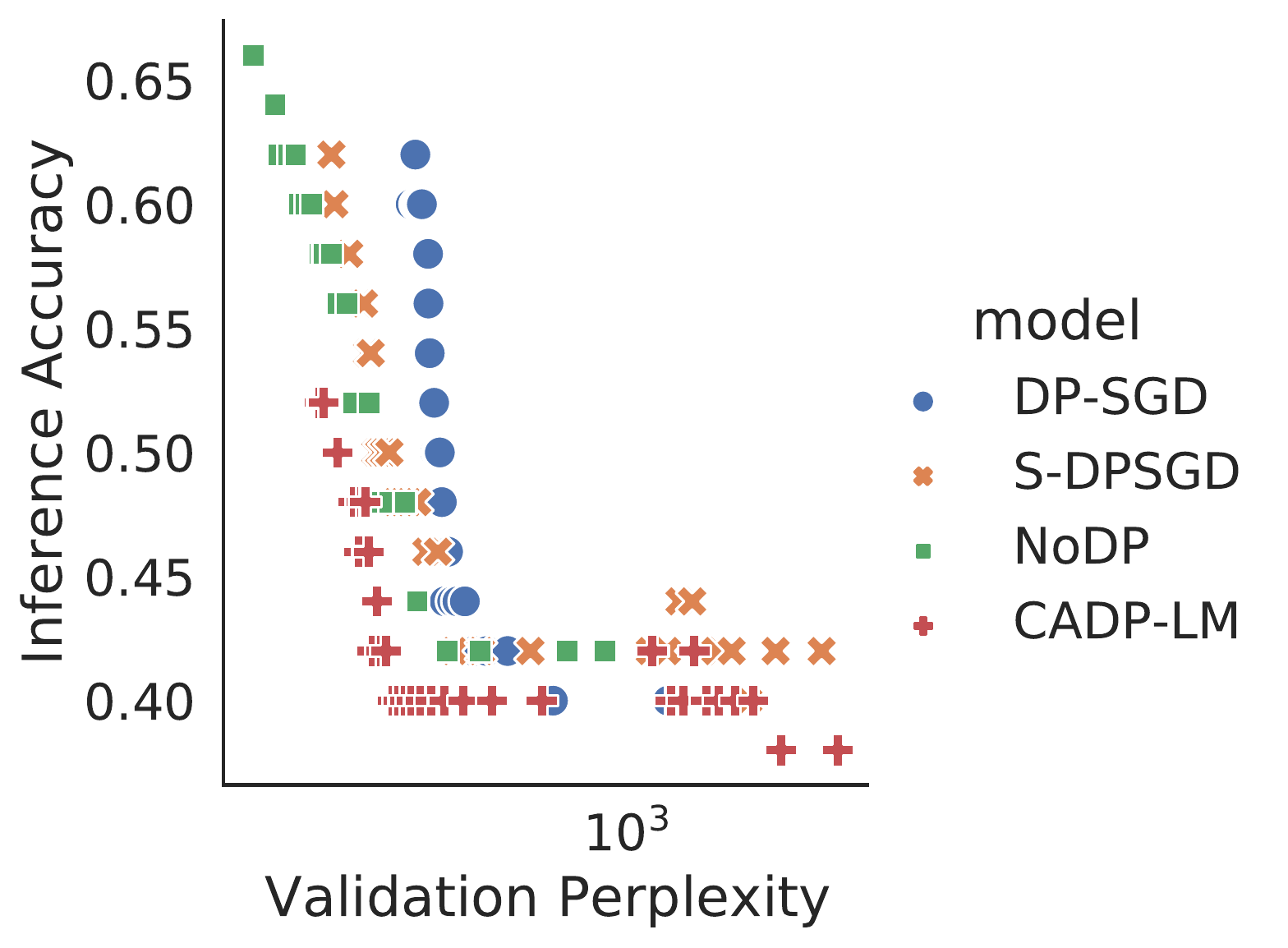}
         \caption{Membership Inference}
        \label{fig:wiki_memb}
     \end{subfigure}
          \begin{subfigure}[b]{0.3\textwidth}
         \centering
         \includegraphics[width=\textwidth]{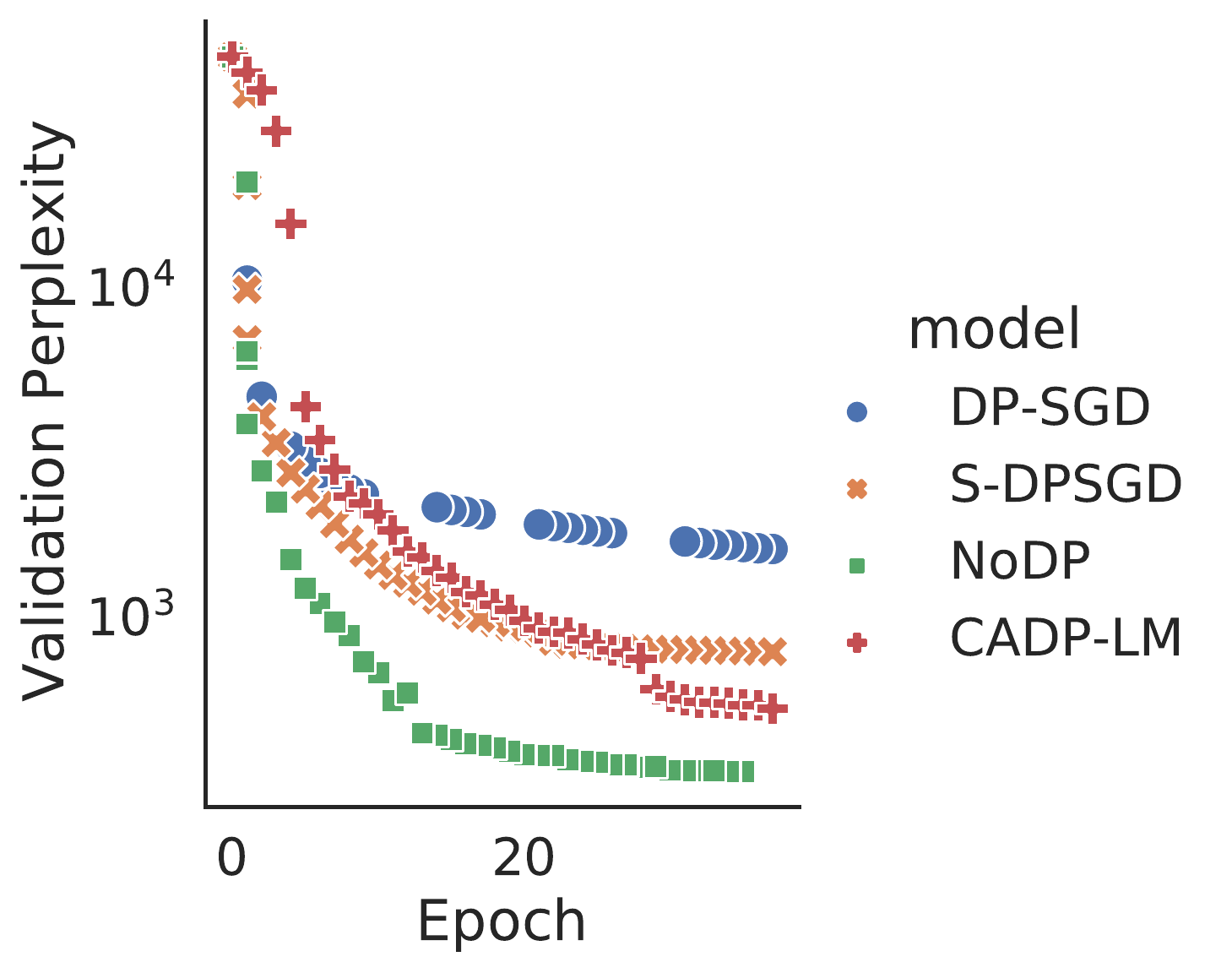}
         \caption{Learning curve}
                 \label{fig:redditacc}

     \end{subfigure}
     \begin{subfigure}[b]{0.3\textwidth}
         \centering
         \includegraphics[width=\textwidth]{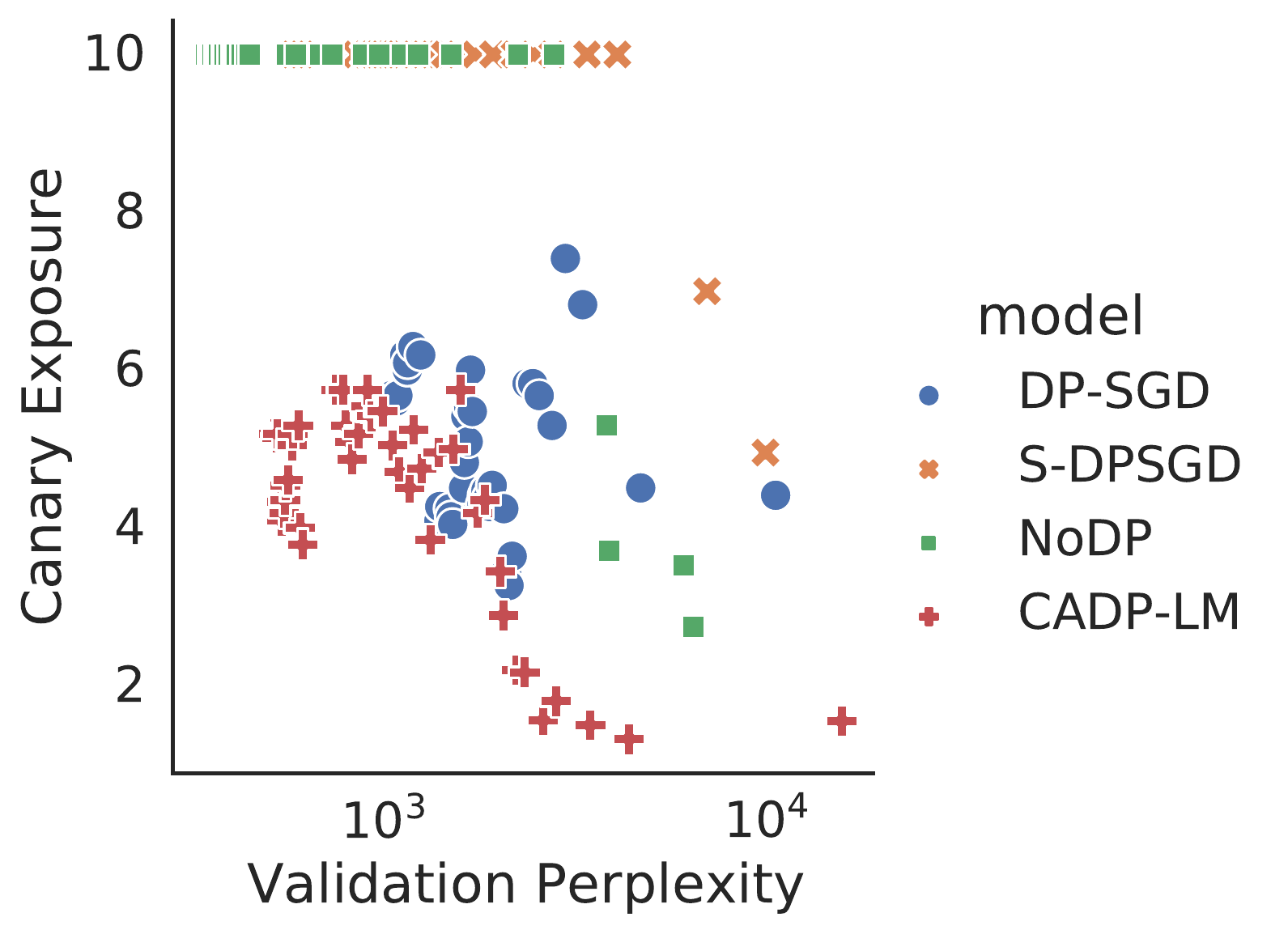}
         \caption{Canary Insertion}
                 \label{fig:reddi_insert}

     \end{subfigure}
     \hfill
     \begin{subfigure}[b]{0.3\textwidth}
         \centering
         \includegraphics[width=\textwidth]{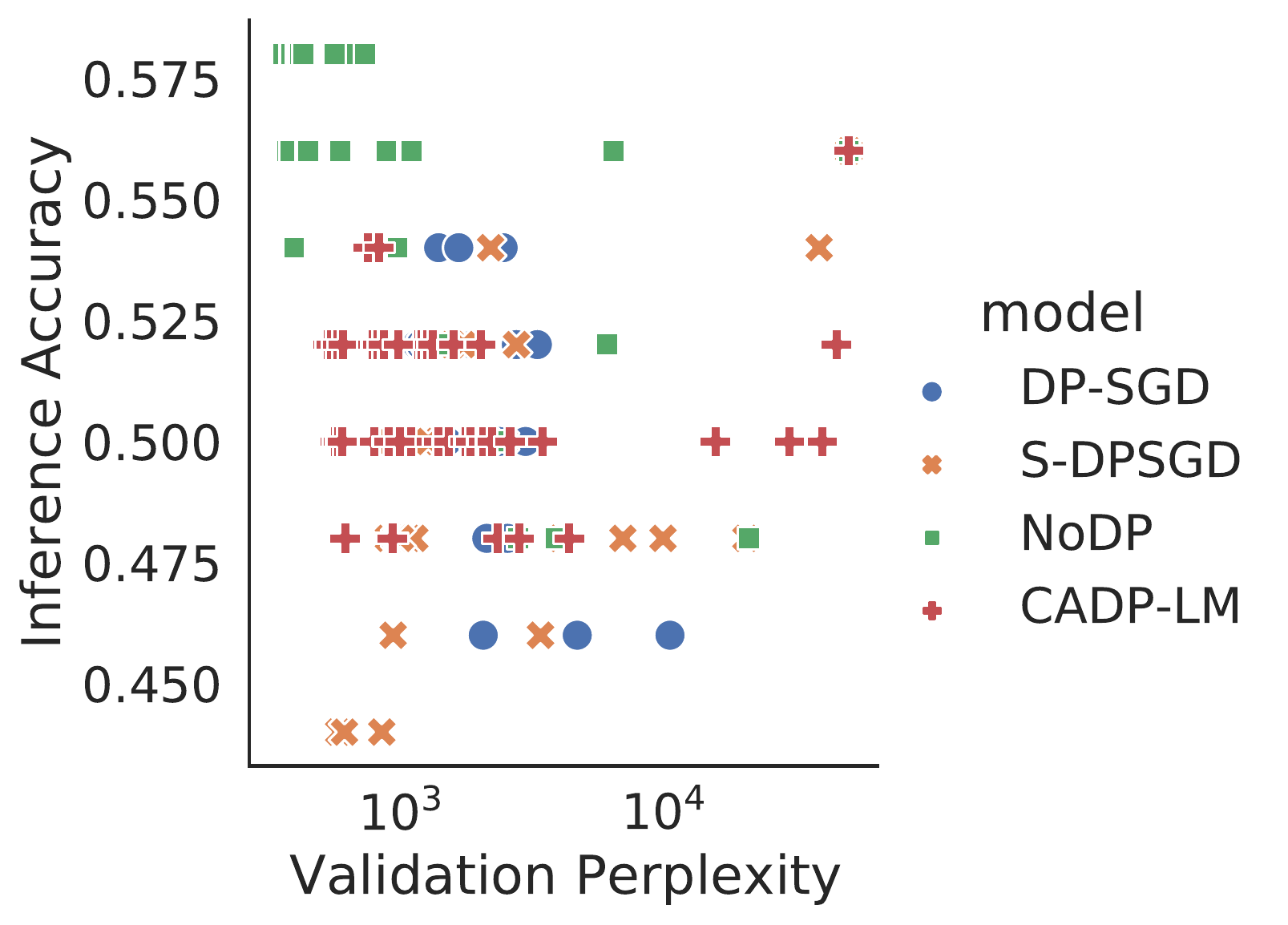}
         \caption{Membership Inference}
                 \label{fig:reddit_memb}

     \end{subfigure}
        \caption{Learning curve, and model's robustness against adversarial attack on WikiText-2 (top), and Reddit Clean Joke dataset (bottom).}

\end{figure}

Figures \ref{fig:wikiacc} and \ref{fig:redditacc} illustrates validation perplexity over epochs. The $y$-axis depicts the model perplexity. Recall that low perplexity is associated with confident model predictions. 
Note how performance degrades when using \textsl{DP-SGD}. In contrast both \textsl{CADP-LM} and \textsl{S-DPSGD}, by adding less noise, are able to retain a higher performance 
We also note that privacy-preserving models trained on \textsl{Reddit}'s data are worse than those trained on \textsl{WikiText-2}, which can explained by the difference in dataset sizes: \textsl{Reddit}'s dataset is about 50 times smaller than \textsl{WikiText-2}. 

Figures \ref{fig:wiki_insert},  \ref{fig:wiki_memb}, \ref{fig:reddi_insert} ,and \ref{fig:reddit_memb} illustrate the effectiveness of privacy preserving mechanisms against canary insertion attack and membership inference attack on \textsl{WikiText-2} (top) and \textsl{Reddit} (bottom). The x-axes report the models’ utilities measured by validation perplexity while the y-axes report the exposure and membership inference accuracy, which indicate the attacks' success. Stronger privacy models have lower exposure score and lower inference accuracy. 

For the canary insertion attack (Figures (b), (e)), as expected, a model trained with no privacy achieves lowest perplexity. However it also attains the highest levels of exposure (score ranging from 8 to 10), indicating that the inserted canary could be easily revealed by an attacker. 
The models trained using \textsl{DP-SGD} offer more protection than those trained using \textsl{S-DPSGD}, as the latter only focuses on protecting "private" tokens. Both algorithms, however, cannot achieve a good tradeoff in protection and accuracy. 
In contrast, \textsl{CADP-LM} shows significant improvements in the exposure risk for comparable perplexities. This is especially observable in Figure 1(b). 

We also note that models trained with S-DPSGD usually return the inserted canary's rank is within top 5, suggesting that an attacker can guess inserted canary with high confidence. The inserted canary's rank querying from DP-SGD models is within the top 50, offering better protections at the cost of a utility degradation. In contrast, the rank queried by \textsl{CADP-LM} are in the top 80-200, suggesting much better privacy preservation while also obtaining lower perplexity scores.

A similar trend is observable for membership inference attack. With no protection, the models are highly vulnerable to these attacks (NoDP). Both S-DPSGD and DP-SGD offer similar protection with S-DPSGD achieving lower perplexity. In contrast, the \textsl{CADP-LM} curve is always below that of both baselines for WikiText-2, suggesting that it outperforms these models both in terms of privacy and utility. This gain in performance is a bit less visible in the Reddit dataset, where, however, it is clear that \textsl{CADP-LM} performs at least as well as the other privacy-preserving baselines. 
{\em The results are significant: By exploiting a notion of context, CADP-LM can achieve superior privacy protection without sacrificing too much the model's utility.}

\section{Conclusions}
This paper was motivated by the rapid adoption of language models (LM) in many inference task for consequential decisions and by adoption of large, often sensitive datasets to train such models. 
We first discussed the current shortcomings of Differential Privacy methods to protect LMs outputs. We then introduced a context-aware DP Language model (CADP-LM) which detect the context in which sensitive information may be revealed and apply a privacy-preserving steps on the predicted sensitive sentences. 
The experimental analysis shows the benefits of this framework on privacy and utility evaluations metrics. 

\section*{Acknowledgments}

This research is partially supported by NSF grant 2133169
and NSF CAREER Award 2143706. Fioretto is also supported by a Google Research Scholar Award and an Amazon Research Award. 
Its views and conclusions are those of the authors only. 

\bibliographystyle{abbrvnat}
\bibliography{references}

\begin{thebibliography}{18}
\providecommand{\natexlab}[1]{#1}
\providecommand{\url}[1]{\texttt{#1}}
\expandafter\ifx\csname urlstyle\endcsname\relax
  \providecommand{\doi}[1]{doi: #1}\else
  \providecommand{\doi}{doi: \begingroup \urlstyle{rm}\Url}\fi

\bibitem[Red()]{Redditdataset}
Reddit clean joke dataset.
\newblock URL
  \url{https://www.kaggle.com/datasets/cuddlefish/reddit-rjokes/code}.

\bibitem[Abadi and et~al.(2016)]{abadi:16}
Abadi and et~al.
\newblock Deep learning with differential privacy.
\newblock In \emph{Proceedings of the 2016 ACM SIGSAC Conference on Computer
  and Communications Security}, 2016.

\bibitem[Bagdasaryan et~al.(2019)Bagdasaryan, Poursaeed, and
  Shmatikov]{bagdasaryan}
E.~Bagdasaryan, O.~Poursaeed, and V.~Shmatikov.
\newblock \emph{Differential Privacy Has Disparate Impact on Model Accuracy}.
\newblock Curran Associates Inc., Red Hook, NY, USA, 2019.

\bibitem[Brown et~al.(2022)Brown, Lee, Mireshghallah, Shokri, and
  Tram\`{e}r]{brown}
H.~Brown, K.~Lee, F.~Mireshghallah, R.~Shokri, and F.~Tram\`{e}r.
\newblock What does it mean for a language model to preserve privacy?
\newblock In \emph{2022 ACM Conference on Fairness, Accountability, and
  Transparency}, FAccT '22, page 2280–2292, New York, NY, USA, 2022.
  Association for Computing Machinery.
\newblock ISBN 9781450393522.
\newblock \doi{10.1145/3531146.3534642}.
\newblock URL \url{https://doi.org/10.1145/3531146.3534642}.

\bibitem[Brown et~al.(2020)Brown, Mann, Ryder, Subbiah, Kaplan, Dhariwal,
  Neelakantan, Shyam, Sastry, Askell, Agarwal, Herbert-Voss, Krueger, Henighan,
  Child, Ramesh, Ziegler, Wu, Winter, Hesse, Chen, Sigler, Litwin, Gray, Chess,
  Clark, Berner, McCandlish, Radford, Sutskever, and Amodei]{gpt3}
T.~Brown, B.~Mann, N.~Ryder, M.~Subbiah, J.~D. Kaplan, P.~Dhariwal,
  A.~Neelakantan, P.~Shyam, G.~Sastry, A.~Askell, S.~Agarwal, A.~Herbert-Voss,
  G.~Krueger, T.~Henighan, R.~Child, A.~Ramesh, D.~Ziegler, J.~Wu, C.~Winter,
  C.~Hesse, M.~Chen, E.~Sigler, M.~Litwin, S.~Gray, B.~Chess, J.~Clark,
  C.~Berner, S.~McCandlish, A.~Radford, I.~Sutskever, and D.~Amodei.
\newblock Language models are few-shot learners.
\newblock In H.~Larochelle, M.~Ranzato, R.~Hadsell, M.~Balcan, and H.~Lin,
  editors, \emph{Advances in Neural Information Processing Systems}, volume~33,
  pages 1877--1901. Curran Associates, Inc., 2020.
\newblock URL
  \url{https://proceedings.neurips.cc/paper/2020/file/1457c0d6bfcb4967418bfb8ac142f64a-Paper.pdf}.

\bibitem[Carlini et~al.(2019)Carlini, Liu, Erlingsson, Kos, and
  Song]{Carlini2019}
N.~Carlini, C.~Liu, U.~Erlingsson, J.~Kos, and D.~Song.
\newblock The secret sharer: Evaluating and testing unintended memorization in
  neural networks.
\newblock In \emph{Proceedings of the 28th USENIX Conference on Security
  Symposium}, SEC'19, page 267–284, USA, 2019. USENIX Association.
\newblock ISBN 9781939133069.

\bibitem[Carlini et~al.(2021)Carlini, Tram{\`e}r, Wallace, Jagielski,
  Herbert-Voss, Lee, Roberts, Brown, Song, Erlingsson, Oprea, and
  Raffel]{carlini2020}
N.~Carlini, F.~Tram{\`e}r, E.~Wallace, M.~Jagielski, A.~Herbert-Voss, K.~Lee,
  A.~Roberts, T.~Brown, D.~Song, {\'U}.~Erlingsson, A.~Oprea, and C.~Raffel.
\newblock Extracting training data from large language models.
\newblock In \emph{30th USENIX Security Symposium (USENIX Security 21)}, pages
  2633--2650. USENIX Association, Aug. 2021.
\newblock ISBN 978-1-939133-24-3.
\newblock URL
  \url{https://www.usenix.org/conference/usenixsecurity21/presentation/carlini-extracting}.

\bibitem[Dwork et~al.(2014)Dwork, Roth, et~al.]{dwork:14}
C.~Dwork, A.~Roth, et~al.
\newblock The algorithmic foundations of differential privacy.
\newblock \emph{Foundations and Trends{\textregistered} in Theoretical Computer
  Science}, 9\penalty0 (3--4):\penalty0 211--407, 2014.

\bibitem[Feldman(2020)]{vitaly}
V.~Feldman.
\newblock Does learning require memorization? a short tale about a long tail.
\newblock In \emph{Proceedings of the 52nd Annual ACM SIGACT Symposium on
  Theory of Computing}, STOC 2020, page 954–959, New York, NY, USA, 2020.
  Association for Computing Machinery.
\newblock ISBN 9781450369794.
\newblock \doi{10.1145/3357713.3384290}.
\newblock URL \url{https://doi.org/10.1145/3357713.3384290}.

\bibitem[Inan et~al.(2021)Inan, Ramadan, Wutschitz, Jones, Rühle, Withers, and
  Sim]{inan}
H.~A. Inan, O.~Ramadan, L.~Wutschitz, D.~Jones, V.~Rühle, J.~Withers, and
  R.~Sim.
\newblock Training data leakage analysis in language models, 2021.
\newblock URL \url{https://arxiv.org/abs/2101.05405}.

\bibitem[Jiang et~al.(2021)Jiang, Araki, Ding, and
  Neubig]{jiang-etal-2021-know}
Z.~Jiang, J.~Araki, H.~Ding, and G.~Neubig.
\newblock How can we know when language models know? on the calibration of
  language models for question answering.
\newblock \emph{Transactions of the Association for Computational Linguistics},
  9:\penalty0 962--977, 2021.
\newblock \doi{10.1162/tacl_a_00407}.
\newblock URL \url{https://aclanthology.org/2021.tacl-1.57}.

\bibitem[Kerrigan et~al.(2020)Kerrigan, Slack, and
  Tuyls]{kerrigan-etal-2020-differentially}
G.~Kerrigan, D.~Slack, and J.~Tuyls.
\newblock Differentially private language models benefit from public
  pre-training.
\newblock In \emph{Proceedings of the Second Workshop on Privacy in NLP}, pages
  39--45, Online, Nov. 2020. Association for Computational Linguistics.
\newblock \doi{10.18653/v1/2020.privatenlp-1.5}.
\newblock URL \url{https://aclanthology.org/2020.privatenlp-1.5}.

\bibitem[Merity et~al.(2016)Merity, Xiong, Bradbury, and
  Socher]{merity2016pointer}
S.~Merity, C.~Xiong, J.~Bradbury, and R.~Socher.
\newblock Pointer sentinel mixture models.
\newblock \emph{arXiv preprint arXiv:1609.07843}, 2016.

\bibitem[Radford et~al.(2019)Radford, Wu, Child, Luan, Amodei, and
  Sutskever]{radford2019language}
A.~Radford, J.~Wu, R.~Child, D.~Luan, D.~Amodei, and I.~Sutskever.
\newblock Language models are unsupervised multitask learners.
\newblock 2019.

\bibitem[Raffel et~al.(2019)Raffel, Shazeer, Roberts, Lee, Narang, Matena,
  Zhou, Li, and Liu]{https://doi.org/10.48550/arxiv.1910.10683}
C.~Raffel, N.~Shazeer, A.~Roberts, K.~Lee, S.~Narang, M.~Matena, Y.~Zhou,
  W.~Li, and P.~J. Liu.
\newblock Exploring the limits of transfer learning with a unified text-to-text
  transformer, 2019.
\newblock URL \url{https://arxiv.org/abs/1910.10683}.

\bibitem[Ramaswamy et~al.(2020)Ramaswamy, Thakkar, Mathews, Andrew, McMahan,
  and Beaufays]{Ramaswamy}
S.~Ramaswamy, O.~Thakkar, R.~Mathews, G.~Andrew, H.~B. McMahan, and
  F.~Beaufays.
\newblock Training production language models without memorizing user data,
  2020.
\newblock URL \url{https://arxiv.org/abs/2009.10031}.

\bibitem[Sanh et~al.(2019)Sanh, Debut, Chaumond, and
  Wolf]{https://doi.org/10.48550/arxiv.1910.01108}
V.~Sanh, L.~Debut, J.~Chaumond, and T.~Wolf.
\newblock Distilbert, a distilled version of bert: smaller, faster, cheaper and
  lighter, 2019.
\newblock URL \url{https://arxiv.org/abs/1910.01108}.

\bibitem[Shi et~al.(2022)Shi, Cui, Li, Jia, and Yu]{shi-etal-2022-selective}
W.~Shi, A.~Cui, E.~Li, R.~Jia, and Z.~Yu.
\newblock Selective differential privacy for language modeling.
\newblock In \emph{Proceedings of the 2022 Conference of the North American
  Chapter of the Association for Computational Linguistics: Human Language
  Technologies}, pages 2848--2859, Seattle, United States, July 2022.
  Association for Computational Linguistics.
\newblock \doi{10.18653/v1/2022.naacl-main.205}.
\newblock URL \url{https://aclanthology.org/2022.naacl-main.205}.

\end{thebibliography}



\end{document}